**Retrieval Augmented Generation for Topic Modeling in Organizational Research:
An Introduction with Empirical Demonstration**

Gerion Spielberger[1], Florian M. Artinger[2], Jochen Reb[3] and Rudolf Kerschreiter[1]

[1]Division of Social, Organizational, and Economic Psychology, Freie Universität Berlin, Germany

[2]Berlin International University of Applied Sciences, Germany

[3]Lee Kong Chian School of Business, Singapore Management University, Singapore

**Author Note**

Gerion Spielberger
Email: gerion.spielberger@fu-berlin.de

Florian M. Artinger
https://orcid.org/0000-0001-9572-2329
Email: artinger@berlin-international.de

Jochen Reb
https://orcid.org/0000-0003-0233-8016
Email: jreb@smu.edu.sg

Rudolf Kerschreiter
https://orcid.org/0000-0003-0281-0261
Email: rudolf.kerschreiter@fu-berlin.de

**Abstract**

Analyzing textual data is the cornerstone of qualitative research. While traditional methods such as grounded theory and content analysis are widely used, they are labor-intensive and time-consuming. Topic modeling offers an automated complement. Yet, existing approaches, including LLM-based topic modeling, still struggle with issues such as high data preprocessing requirements, interpretability, and reliability. This paper introduces Agentic Retrieval-Augmented Generation (Agentic RAG) as a method for topic modeling with LLMs. It integrates three key components: (1) *retrieval*, enabling automatized access to external data beyond an LLM's pre-trained knowledge; (2) *generation*, leveraging LLM capabilities for text synthesis; and (3) *agent-driven learning*, iteratively refining retrieval and query formulation processes. To empirically validate Agentic RAG for topic modeling, we reanalyze a Twitter/X dataset, previously examined by Mu et al. (2024a). Our findings demonstrate that the approach is more efficient, interpretable and at the same time achieves higher reliability and validity in comparison to the standard machine learning approach but also in comparison to LLM prompting for topic modeling. These results highlight Agentic RAG's ability to generate semantically relevant and reproducible topics, positioning it as a robust, scalable, and transparent alternative for AI-driven qualitative research in leadership, managerial, and organizational research.

Keywords: qualitative data analysis, retrieval augmented generation, topic modeling, machine learning

## Introduction

Qualitative research relies heavily on analyzing textual data in the form of interviews, open-ended survey responses, email communication, news, or company documents. Doing such analyses is both time-intensive and resource-demanding, even with the aid of widely-used qualitative data analysis software. One well-established method for qualitative data analysis in managerial research is topic modeling, a *machine learning* approach that identifies relevant topics within textual datasets (Blei et al., 2003; Tonidandel et al., 2022). Topic modeling can be considered a partly automated approach of traditional qualitative data analysis techniques such as grounded theory or content analysis (Baumer et al., 2017; Schmiedel et al., 2019; van der Velde & Gerpott, 2023). While topic modeling has effectively reduced the need for human involvement for coding, the process itself remains labor-intensive, and the resulting topics are often difficult to interpret (Mu et al., 2024a; Schmiedel et al., 2019).

Advancements in *Artificial Intelligence* (AI) and *Large Language Models (LLMs)* suggest significant potential in improving the processing and contextual understanding of textual data, providing promising solutions to these challenges. Indeed, as LLMs continue to evolve, they are increasingly applied to automate traditional text analysis frameworks, including grounded theory (Übellacker, 2024) and topic modeling (Mu et al., 2024a; Mu et al., 2024b). Recent studies demonstrate that LLMs can identify relevant topics in textual datasets with a high alignment to human coding, underlining their potential for qualitative data analysis (Mu et al., 2024a; Mu et al., 2024b). However, significant challenges remain in terms of transparency and leave room for improvements with regards to efficiency, validity, and reliability for text data analysis with the integration of LLMs.

To address these challenges, this paper proposes incorporating Agentic Retrieval-Augmented Generation (Agentic RAG) into the topic modeling process, particularly in the context of social science and organizational research. LLMs are trained on vast amounts of publicly available data. A key advantage of Agentic RAG is that it uses the capabilities of LLMs with respect to handling natural language and applies these capabilities to data that the LLM was not previously trained on. RAG enables access to up-to-date, domain-specific, and proprietary information, including research databases, interview transcripts, and other specialized datasets. In other words, it 'augments' the knowledge base of the LLM from which it 'retrieves' information.

In this article, we argue that, through this powerful "global-local" combination, together with the integration of *LLM agents* for better retrieval quality, Agentic RAG can substantially improve topic modeling with regards to four dimensions: (1) efficiency, (2) transparency, (3) validity, and (4) reliability. We do so by laying out the limitations of present approaches, explaining the benefits of Agentic RAG and testing it on an empirical dataset.

Overall, this research makes two key contributions. First, our study introduces Agentic RAG to organizational scholarship and elaborates how it improves topic modeling compared to both traditional methods and LLM-based prompting approaches, with a particular focus on the dimensions of efficiency and transparency. Second, the study validates the functionality of Agentic RAG by empirically assessing its validity and reliability, providing measurable evidence of its effectiveness.

## Theoretical Background

**Traditional Textual Data Analysis**

Qualitative data analysis holds significant importance in organizational research, such as in understanding leadership behavior through qualitative interviews, the analysis of voice or video-recordings of leader-employee interactions through multimodal LLMs that can analyze not only text, but also sounds or images (Mu, H. et al., 2024). For simplicity, our focus here will be text, including text that is transcribed from audio data. Traditionally, qualitative data analysis has been dominated by two primary methodological approaches: grounded theory and content analysis. While case study research (Eisenhardt, 1989; Pratt, 2025) provides a valuable framework for contextual analysis and theory development, our emphasis is on methodological approaches for analyzing qualitative data rather than on case-based research designs. Grounded theory and content analysis, in particular, serve as the foundation for many qualitative studies, offering systematic methods for identifying themes and patterns in textual data. Both approaches employ manual coding and remain widely utilized in contemporary research. Gioia et al. (2013) systematized the grounded theory approach within the context of organizational studies, integrating it into the inductive research process for qualitative data analysis. The primary objective of this approach is to develop a theory grounded in the data. In contrast, content analysis aims to address specific research questions by identifying and analyzing themes or patterns within textual data (Duriau et al., 2007). Grounded theory and content analysis have been enhanced by computational tools in recent years. Software such as MAXQDA and other computer-assisted qualitative data analysis software have facilitated the structuring of the research process and the performance of quantitative analyses, such as word distributions (O'Kane et al., 2021). Both methods involve time-consuming processes for coding and interpretation, requiring substantial analytical effort and resources.

**Topic Modeling**

In recent years, topic modeling has emerged as an automated method for content analysis and grounded theory, offering advantages such as reduced coding effort, the capacity to analyze larger datasets and increased objectivity (Baumer et al., 2017; Blei et al., 2003; Schmiedel et al., 2019; van der Velde & Gerpott, 2023). Topic modeling is a machine learning approach that identifies the most relevant topics within a dataset using statistical algorithms. Researchers can contextualize and interpret the results to generate a theory from the data or answer a specific research question. Topic modeling has gained increasing popularity in organizational research (Schmiedel et al., 2019; Tonidandel et al., 2022; van der Velde & Gerpott, 2023).

In the following section we discuss three methods for topic modeling: (1) the traditional method with Latent Dirichlet Allocation (LDA) as well as (2) using an LLM with prompting. We then introduce (3) Agentic RAG for topic modeling.

## Methods for Topic Modeling

**Traditional Topic Modeling with Latent Dirichlet Allocation (LDA)**

Topic Modeling using the LDA algorithm is one of the most widely used methods (Blei et al., 2003; Schmiedel et al., 2019). It is based on a statistical model that incorporates natural language processing to identify groups of frequently co-occurring words within textual data. Before applying the algorithm to the data, the dataset requires extensive preprocessing. This process typically includes steps such as data cleaning (e.g., *removing stop words*) and text standardization like *lemmatization* and *stemming*. After preprocessing, the algorithm generates lists of subtopics of co-occurring and semantically related terms. These lists must then be analyzed and interpreted by the researcher to assign overarching labels to the main topics.

Although the well-established topic modeling approach can aid the textual data analysis process by reducing resources for human coding, it has several limitations. First, the topic modeling process using LDA remains resource-intense, requires data cleaning, as well as manual coding and interpretation for accurate topic identification (Schmiedel et al., 2019). A second limitation of LDA is its reliance on preprocessing steps, such as stemming and lemmatization, which can significantly influence model performance (Chuang et al., 2015; Schofield & Mimno, 2016). While these steps aim to standardize the data, they can lead to information loss and potential miss of important linguistic nuances that contribute to a richer understanding of the text. Third, the topics generated by LDA frequently lack intuitive alignment with human understanding, making them challenging to interpret without additional contextual information (Gillings & Hardie, 2023; Mu et al., 2024a).

**LLM Prompting for Topic Modeling**

As LLMs become more prevalent and popular in society and research, many researchers are evaluating the potential of LLMs. Recently, studies have been utilizing LLMs for item and scale development (Fyffe et al., 2024), data analysis (Tai et al., 2024), and also topic modeling (Mu et al., 2024a). LLMs demonstrate high efficiency in processing substantial amounts of textual data. In a series of experiments, Mu et al. (2024a) developed an approach whereby LLMs (e.g., GPT and LLaMA) were used to extract relevant topics from extensive datasets to perform topic modeling via prompts. Because of the capacity limitations of the models, the data was provided in subsets and from each subset the most relevant topics were identified by the LLM. In a final step, all identified topics from every chunk were summarized with another prompt to a final number of ten main topics. Because the data is processed based on the formulation of the prompt, we refer to this approach as LLM prompting.

Their findings suggest that LLMs generate topics more aligned with human coding than those produced by methods like LDA. The performance of the LLM was strong even when applied to datasets with limited context, such as short tweets. This demonstrates a clear advantage over LDA, which often struggles to extract meaningful insights from short or highly condensed text data (Laureate et al., 2023).

These results highlight the potential of LLMs for research tasks such as topic modeling, offering significant time savings in human data processing while maintaining high-quality results. However, several critical limitations need to be discussed. One limitation is the issue of feeding the data in subsets, due to LLM capacities. This is not only inefficient but also prone to the risk of errors. Multiple steps of feeding data as well as summarizing the generated subtopics topics are needed. Moreover, retrieval using LLMs presents additional challenges, such as hallucinations and a lack of transparency, making the process a "black box". Hallucinations occur when LLMs generate information that seems plausible but is incorrect, fabricated, or unsupported by any data. It stems on the one hand from the inherent stochasticity of the LLMs where words or sequences of words are predicted probabilistically – by design some predictions will fail. On the other hand, hallucinations can stem from insufficient training data, whereby the topic that the user queries is insufficiently covered; a problem which is heightened by feeding the data in subsets. Hallucinations are one of the major issues of LLMs and generative AI in general.

As reported in the study of Mu et al. (2024a) the authors experimented with multiple prompt variations and even adding seed topics (example topics as an anchor for the LLM), highlighting the significant impact of prompt formulation on output quality. This reliance on precise prompt design brings complexity and unpredictability into the process and may hamper reproducibility. A third critical issue is the lack of transparency. While LLMs are highly capable, many of them provide – at least at present – limited insight into how the output is generated. Consequently, in most cases it remains unclear whether the output represents the best possible result or merely the first sufficient one identified by the LLM. More generally, at present it is often not possible to reconstruct how the model arrived at its output. Particularly in a scientific context, in which it is central to understand the chain of thought that leads to a particular result, such a limitation may be troublesome.

**Agentic RAG in Textual Data Analysis**

To address limitations of current LLM-based approaches for topic modeling, we propose an enhanced approach: agentic Retrieval-Augmented Generation (Agentic RAG). Agentic RAG extents traditional RAG by incorporating specialized LLM agents to enhance the data processing.

RAG, first introduced by Lewis et al., (2021) is a framework to enlarge the knowledge base of an LLM by integrating external data sources. This approach enables interaction with the data through *queries* (similar to prompts). By integrating an external retriever, RAG identifies and retrieves data relevant to the query. This retrieved information from the data is combined with the query and contextual knowledge, allowing the LLM to generate an informed response. The integration of LLM agents (in our case the ReAct agent) for Agentic RAG further refines the process by incorporating evaluation and reformulation mechanisms. If the retrieved information does not meet the evaluation criteria, the query is reformulated, and another retrieval iteration is initiated. This iterative process continues until the question is sufficiently answered, thereby optimizing the overall workflow.

**From Naïve to Agentic RAG**

The RAG framework is designed to enhance the capabilities of a LLM by making external data accessible to it. LLMs are trained on vast amounts of data, enabling them to perform impressively across a wide range of tasks and topics. However, their knowledge is inherently limited to the public data. Consequently, if we seek information on events or developments that are proprietary like research data, the LLM cannot provide accurate or up-to-date responses.

The most basic form of RAG is referred to as "Naïve RAG" as it consists only of the basic building blocks, illustrated in Figure 1. In the first step of the process, the original data, such as documents or text, is split into smaller segments called "chunks," which may consist of a few sentences or a short paragraph, depending on the chunking strategy used. Each chunk is then encoded into a numerical vector (*embedding*) using a pre-trained model that captures the semantic meaning of the text. These vectors are stored as documents in a vector database optimized for searching based on similarity and serves as the knowledge base. When a researcher poses a question in form of a query, the LLM transforms the query into a vector, which is then compared against the stored vectors in the knowledge base using a retriever model to find the most semantically relevant documents. The most relevant documents are retrieved and, together with the initial query, fed to the LLM, which uses them as input to generate a coherent response, synthesizing the retrieved information with its internal knowledge to provide an accurate answer to the query. While RAG enhances the retrieval and generation process, its performance remains constrained by its linear and static design (Singh et al., 2025). To achieve greater adaptability and performance tailored to specific needs, it can be further advanced through the integration of LLM agents.

**Figure 1**

*Illustration of the key components of Naïve RAG*

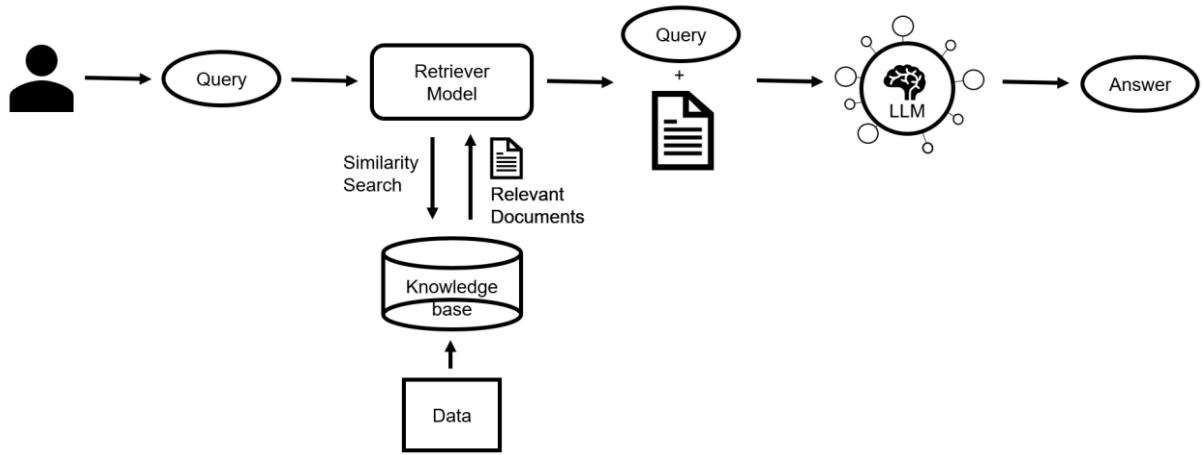

**Agentic RAG**

The RAG framework allows potential for advancement at each key step of the RAG process, including chunking, embedding, retrieval, and prompting. Users can customize the process for their specific needs. It also allows the integration of LLM agents for better retrieval quality. LLM agents are intelligent entities that can execute specific tasks autonomously (Singh et al., 2025). In our approach, we use the ReAct agent, which operates on two principles: "Reason" and "Act" (Yao et al., 2023). The agent requires the LLM to explicitly articulate its "reasoning" process, followed by a description of the subsequent action.

The tasks of the ReAct agent involve assessing generated outputs, reformulating queries and re-retrieving data as needed to optimize results using the ReAct agent (Yao et al., 2023). In each retrieval cycle, results are analyzed, and if they do not meet the criteria —such as failing to fully address the initial query, lacking factual consistency, containing ambiguous or conflicting information—, the agent reformulates the query and initiates another iteration, followed by further result analysis. This iterative process offers several advantages. First, it reduces dependency on a perfectly crafted initial query, allowing for more flexible, dynamic adjustments based on retrieval outcomes. If specific criteria are required for query reformulation, they can be predefined to ensure consistency. Second, the agent's iterative reassessment avoids premature satisfaction with early results, ensuring that the retrieved output is not simply the first plausible option but the most contextually aligned and accurate result. These "Reasoning" and "Acting" functions of the agent are textualized in the generated output, making it more transparent for the researcher. Figure 2 illustrates an excerpt of the ReAct agent's "reasoning" and "acting" process during topic identification. The example demonstrates two distinct steps in the agent's workflow: In the first step the agent first formulates a query to retrieve relevant documents from the knowledge base. It strategically uses diverse phrasings to ensure a comprehensive data search. In the second step, after retrieving relevant documents, the agent evaluates the data and extracts key topics related to vaccine hesitancy, ensuring each topic remains concise and relevant. By capturing the agent's "thoughts", "actions", and "reasoning", Figure 2 highlights how Agentic RAG enhances transparency and provides researchers with insight into the LLM's decision-making process. This structured approach ensures that the output is not only data-driven but also interpretable, reducing the "black-box" nature of LLM-generated results.

**Figure 2**

*Example of the ReAct Agents "thought" process in action*

```
=== Agent thoughts:
Thought: To identify topics related to COVID-19 vaccine hesitancy, I will
perform multiple queries using the 'retriever' tool. I will use diverse
phrasings to ensure a comprehensive exploration of the data.
>>> Calling tool: 'retriever' with arguments: {'query': 'Identify specific
topics related to COVID-19 vaccine hesitancy.'}
Retrieved documents: (…)
Thought: Based on the retrieved documents, I will now identify specific
topics related to COVID-19 vaccine hesitancy, ensuring each topic is no
more than three words.

Action:
{
  "action": "final_answer",
  "action_input": {
    "answer": [
      "Topic 1: Risk-Benefit Analysis",
      "Topic 2: Vaccine Safety",
      "Topic 3: Long-Term Effects",
      "Topic 4: Natural Immunity",
      "Topic 5: Political Motivation",
      "Topic 6: Vaccine Efficacy",
      "Topic 7: Vaccine Side Effects",
      "Topic 8: Vaccine Misinformation",
      "Topic 9: Vaccine Trust",
      "Topic 10: Vaccine Mandates"
    ]
  }
}
===== Extracting action =====
=== Agent thoughts:
Thought: Based on the retrieved documents, I will now identify specific
topics related to COVID-19 vaccine hesitancy, ensuring each topic is no
more than three words.
>>> Calling tool: 'final_answer' with arguments: {'answer': ['Topic 1:
Risk-Benefit Analysis', 'Topic 2: Vaccine Safety', 'Topic 3: Long-Term
Effects', 'Topic 4: Natural Immunity', 'Topic 5: Political Motivation',
'Topic 6: Vaccine Efficacy', 'Topic 7: Vaccine Side Effects', 'Topic 8:
Vaccine Misinformation', 'Topic 9: Vaccine Trust', 'Topic 10: Vaccine
Mandates']}
```

**Benefits of Agentic RAG for Topic Modeling**

By enlarging the knowledgebase of LLMs with external data and maximizing the quality of information retrieval together with the capabilities of LLMs, Agentic RAG matches the requirements for topic modeling and also mitigates the limitations of former approaches like LDA or LLM prompting as employed in Mu et al. (2024a). We propose two qualitative advantages of Agentic RAG compared to LLM prompting: efficiency and transparency.

Even very capable LLMs (e.g., GPT-4/4o) are constrained by a limit of how much text one can provide, which is quickly reached when processing substantial amounts of data simultaneously. Once this limit is exceeded, the model generates an error. In their study, Mu et al. (2024a) encountered this limitation and addressed it by providing only subsets of data to the LLM. The model then identified the most relevant topics within each subset which were subsequently summarized in a later step. Agentic RAG, in contrast, processes data as numerical embeddings. This approach significantly reduces computational costs and allows for the inclusion of much larger datasets before reaching the text limit. Consequently, Agentic RAG can process an entire dataset, eliminating the need for an additional summarization step. This enhances efficiency by streamlining the process while reducing overall costs.

As detailed above, Agentic RAG enhances transparency by providing insight into the model's "thought processes" through the ReAct Agent making the LLMs "reasoning" and "acting" processes visible. This also increases error detection and makes the system and therefore the quality of the results more comprehendible. Researchers can gain a clearer understanding of the output generation process by employing LLM agents that require the model to evaluate outputs and automatically enhance retrieval through query reformulation.

**Quantitative Evaluation of Agentic RAG for Topic Modeling**

Beyond improving efficiency and transparency, Agentic RAG has the potential to enhance two critical aspects of topic modeling: validity and reliability. To systematically assess these dimensions, we outline the rationale behind our evaluation criteria before presenting the empirical validation.

First, we argue that validity is improved because the Agentic RAG framework ensures that the LLM generates results exclusively based on the provided dataset, rather than relying on its pre-trained knowledge base. This structure is reducing the model's susceptibility to hallucinations (Lewis et al., 2021). Furthermore, the iterative evaluation, reformulation and re-retrieval processes implemented by the agent ensure that the model incorporates all relevant information, rather than stopping at the first plausible result. This structured approach minimizes the risk of irrelevant or incomplete topic generation, enhancing the overall validity of the extracted topics.

Second, reliability is strengthened through an iterative retrieval and query reformulation process facilitated by the ReAct agent. By continually refining and retrieving only the most relevant segments, this approach ensures that the model consistently generates similar topics when applied repeatedly to the same dataset. Moreover, since retrieval is anchored in an external knowledge base rather than solely dependent on a stochastic generative model, the topics remain more stable across runs, reducing variability and increasing methodological robustness.

These advantages suggest that Agentic RAG strengthens both the accuracy and stability of topic modeling, positioning it as a robust alternative to LDA and standard LLM-based prompting. To empirically assess these claims, we conducted a systematic evaluation of Agentic RAG, performing topic modeling on a publicly available dataset and assessing validity and reliability using quantitative metrics.

## Methods

The code implementation of Agentic RAG in this study is inspired by the Hugging Face Agentic RAG framework (https://huggingface.co/learn/cookbook/agent_rag). For the empirical evaluation, we used OpenAIs GPT-4o (https://openai.com/) as the underlying language model, ensuring state-of-the-art reasoning and retrieval capabilities. To promote transparency and reproducibility, the complete codebase is publicly available at (https://github.com/GerionGit/Agentic-RAG).

### Dataset

For empirical validation, we conducted an Agentic RAG topic modeling analysis on the dataset used by Mu et al. (2024a), enabling a direct comparison of performance with their results. Mu et al. (2024a) analyzed two publicly accessible datasets using their LLM-based topic modeling approach, demonstrating the capability of LLMs to perform topic modeling effectively compared to traditional methods such as LDA. To assess the potential improvements offered by Agentic RAG, we focused on the VAXX dataset, comparing our results to those presented by Mu et al. (2024a). The VAXX dataset, originally compiled by Poddar et al. (2022), consists of Twitter data (tweets) related to COVID-19 vaccine hesitancy. To ensure comparability, we used the dataset provided by Mu et al. (2024a) as a CSV file. The data was processed using the Agentic RAG workflow we developed, incorporating modifications to align with the methodological steps outlined by Mu et al. (2024a).

To evaluate the consistency and therefore reliability of the identified topics, we repeated the process of extracting the 10 most relevant topics using Agentic RAG five times. This iterative approach allowed us to assess the reliability of the results. The outcomes of our analysis are summarized in Table 1.

**Measures**

To evaluate both the validity and reliability of our results, we employ cosine similarity, a widely used metric in natural language processing. Cosine similarity measures the degree of similarity between two vectors in a multidimensional space. Within the Agentic RAG framework, our data is represented as vectors encoding the semantic meaning of words. Thus, cosine similarity provides a suitable method for assessing the similarity between topics as well as the overall alignment between the generated outputs and the dataset.

To evaluate validity of the generated topics, we calculate a relevance score between the generated topics and the dataset with cosine similarity. This score reflects the relevance of the topic list in relation to the dataset, providing a weighted measure of how well the topics capture the underlying content. We extend this analysis by applying the same approach to topic lists generated by other methods, such as the model presented in Mu et al. (2024a) and traditional LDA, allowing for a comparative evaluation of topic relevance across different methodologies. The key steps in this process are summarized in Table 1.

**Table 1**

*Evaluating validity*

| Step | Description |
| --- | --- |
| 1. Encoding Topics as Embeddings | A set of top topics was selected for each method (e.g., Round 1 from Agentic RAG, LLaMA Expt. 3 + Summarization from Mu et al., 2024a). These topics were encoded into vector representations using a pre-trained SentenceTransformer model (*all-MiniLM-L6-v2*). |
| 2. Reverse Retrieval of Documents | For each topic, a similarity search was conducted against a vectorized knowledge base to retrieve the most relevant documents. This step leveraged the similarity search functionality of the retriever's vector database. |
| 3. Cosine Similarity Calculation | The cosine similarity between the vector representation of each topic and the embeddings of the retrieved documents was calculated, quantifying the semantic alignment between topics and dataset content. |
| 4. Weighted Average Calculation | To ensure that more frequently occurring topics had a proportionate influence, a weighted average relevance score was computed based on the number of retrieved documents per topic. |
| 5. Comparison Across Methods | The entire process was repeated for each topic modeling method (basic prompting, LDA, and Agentic RAG), enabling a direct comparison of topic relevance scores. |

To evaluate reliability, we calculate the consistency of generated topics across five independent rounds of topic modeling with Agentic RAG. High consistency between rounds indicates reliable topic generation, as it demonstrates the method's stability under similar conditions. Table 2 summarizes the key steps in this evaluation.

**Table 2**

*Evaluating reliability*

| Step | Description |
| --- | --- |
| 1. Topic Generation Across 5 Rounds | Agentic RAG was executed independently five times, each iteration producing a set of 10 topics based on the dataset. |
| 2. Embedding of Topics | Topics from each round were transformed into dense vector embeddings using a pre-trained SentenceTransformer model (*all-MiniLM-L6-v2*), enabling semantic comparisons. |
| 3. Cosine Similarity Calculation | Pairwise cosine similarity matrices were computed for each pair of rounds to quantify the semantic overlap between generated topics. |
| 4. Average Highest Similarity | For each pair of rounds, the highest cosine similarity for each topic in the first round was identified. The average of these maximum values was then calculated to assess topic alignment across iterations. |
| 5. Reliability Metric | The mean of the highest similarity scores between Round 1 and subsequent rounds served as the reliability metric. Higher values indicate greater consistency in topic generation. |

## Results

To examine validity and reliability, we conducted five rounds of topic modeling using Agentic RAG on the COVID-19 vaccine hesitancy dataset, which was previously analyzed in the study by Mu et al. (2024a). Each round generated a set of 10 topics. The extracted topics from all five rounds are presented in Table 3. At first glance, the topics appear highly consistent across the five rounds, with minor variations in wording and categorization (e.g., Political Influence vs. Political Motivation; Health Bureaucracy vs. Vaccine Transparency), suggesting a stable and replicable topic structure.

**Table 3**

*Topics generated in five rounds of topic modeling with Agentic RAG*

| Topic | Round 1 | Round 2 | Round 3 | Round 4 | Round 5 |
|---|---|---|---|---|---|
| 1 | Vaccine Safety | Vaccine Safety | Vaccine Safety | Vaccine Safety | Vaccine Safety |
| 2 | Side Effects | Side Effects | Side Effects | Side Effects | Side Effects |
| 3 | Trust Issues | Trust Issues | Trust Issues | Trust Issues | Trust in Government |
| 4 | Political Influence | Political Influence | Political Motivation | Political Influence | Political Motivation |
| 5 | Long-term Effects | Long-term Effects | Long-term Effects | Long-term Effects | Long-Term Effects |
| 6 | Efficacy Doubts | Efficacy Concerns | Efficacy Concerns | Efficacy Doubts | Vaccine Efficacy |
| 7 | Risk Perception | Risk Perception | Risk-Benefit Analysis | Risk-Benefit Analysis | Risk-Benefit Analysis |
| 8 | Mandatory Vaccination | Mandatory Vaccination | Control Concerns | Mandatory Vaccination | Mandatory Vaccination |
| 9 | Health Bureaucracy | Vaccine Transparency | Vaccine Necessity | Transparency Concerns | Vaccine Development Speed |
| 10 | Conspiracy Theories | Natural Immunity | Natural Immunity | Conspiracy Theories | Natural Immunity |

To quantify the validity of topics generated by Agentic RAG, we calculated weighted cosine similarity between the extracted topics in each method and the full dataset. This approach measures how closely the extracted topics align with the semantic structure of the original data. The results, presented in Figure 3, indicate that Agentic RAG achieves a higher weighted relevance score (0.43) compared to both LLM prompting (0.33) and traditional LDA (0.27) both from the evaluation of Mu et al. (2024a). These findings suggest that Agentic RAG generates topics that are more aligned with the dataset than those produced by alternative methods.

**Figure 3**

*Evaluating Validity across Topic Modeling Methods*

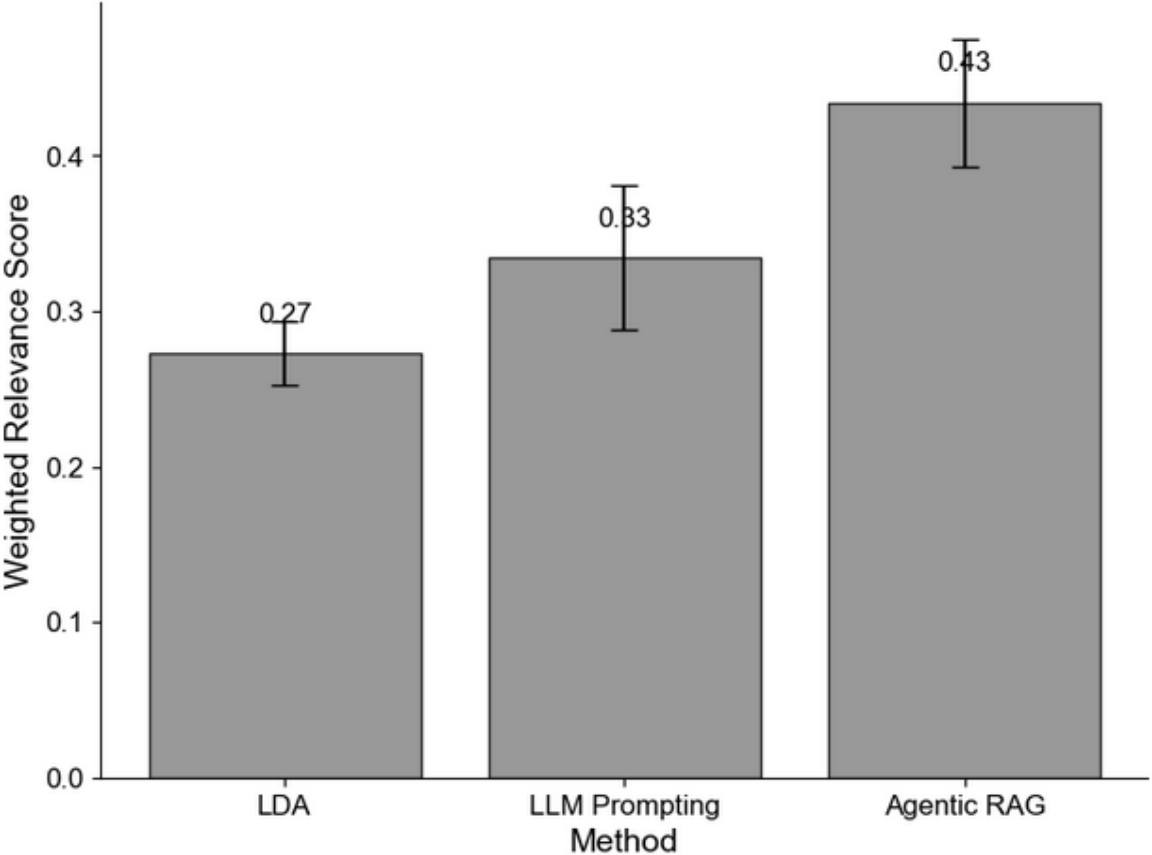

*Note.* Comparison of weighted relevance scores for three topic modeling methods: LLM Prompting Mu et al. (2024a), Latent Dirichlet Allocation (LDA), and Agentic RAG. Bars represent the mean weighted relevance scores across topics, with error bars indicating standard errors (+/-1 SE).

To quantify reliability, we measured topic similarity scores between Round 1 and each subsequent round using cosine similarity. Higher similarity scores indicate greater reliability across iterations, demonstrating the method's stability. As shown in Figure 4, similarity scores range from 0.71 to 0.90, with the highest stability observed between Round 1 and Round 4 (0.90) and the lowest between Round 1 and Round 5 (0.71).

**Figure 4**

*Evaluating Reliability of Agentic RAG across five Iterations of Topic Modeling*

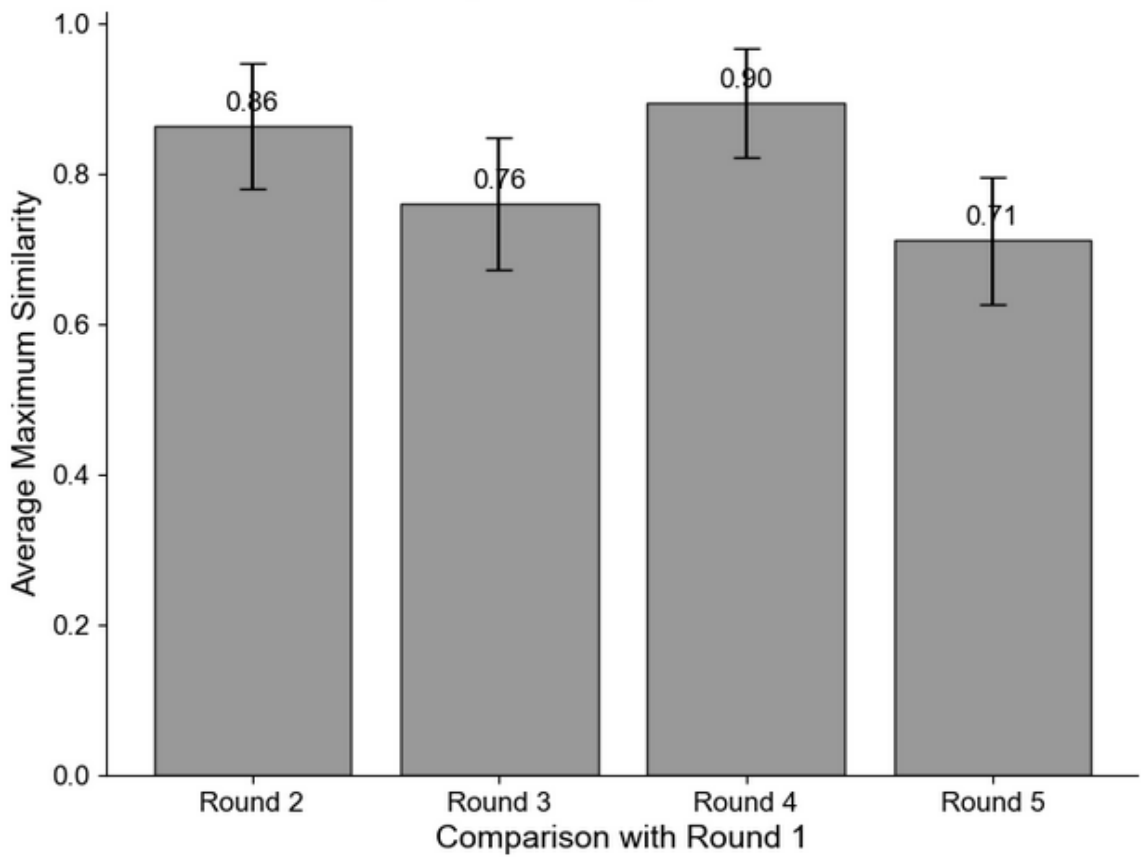

*Note.* Average similarity scores across rounds compared to Round 1, assessing the reliability of Agentic RAG topic modeling results. Bars represent the mean similarity scores, with error bars indicating standard errors. Higher values suggest greater consistency of topic representations across iterations.

Overall, our empirical evaluation demonstrates that Agentic RAG outperforms both LDA and basic LLM prompting in terms of topic validity, generating topics that are stronger aligned to the dataset, indicated by higher weighted cosine similarity scores. Additionally, the results confirm that Agentic RAG produces consistent topics across multiple iterations, ensuring a high level of reliability in topic modeling outcomes. These findings underscore the effectiveness of Agentic RAG as a robust and scalable alternative to existing topic modeling approaches.

**Discussion**

The primary contribution of this paper is to advance text analysis using LLMs and enhance the qualitative data analysis process in organizational research. By introducing Agentic RAG for topic modeling, we illustrate how this approach addresses key limitations of traditional qualitative methods, such as grounded theory and content analysis, while also overcoming challenges associated with existing LLM-based approaches, such as LLM prompting. At the same time, Agentic RAG retains the strengths of these methodologies, making it a robust alternative. Specifically, this approach enhances four critical dimensions of qualitative data analysis: efficiency, transparency, validity, and reliability.

While existing LLM-based methods for topic modeling reduce human coding resources, our Agentic RAG approach goes further by reducing preprocessing requirements. Instead of feeding raw data subsets to the LLM, the data is represented as embeddings in a vector store, which significantly reduces computational costs and the need for resource-intensive preprocessing and summarization steps. This streamlined workflow enhances overall efficiency by simplifying the interaction between data and the model.

With respect to transparency, the integration of the ReAct LLM agent provides important insights into the "reasoning" and "acting" process of the LLM, making it possible to trace and understand the steps taken during topic generation. Transparency is an important concern in LLM-based methods, particularly for scientific applications where interpretability and reliability are crucial. By addressing this issue, Agentic RAG contributes to increasing the trustworthiness and usability of LLM-based tools in qualitative research.

Our empirical findings highlight the quantifiable strengths of Agentic RAG, particularly in validity and reliability. We show that the generated topics with Agentic RAG were more relevant to the dataset, by scoring higher weighted cosine similarity compared to traditional methods like LDA and LLM prompting. Furthermore, the reliability of Agentic RAG was demonstrated through high consistency scores across multiple iterations of topic generating. This reliability confirms that the generated topics are reproducible, making it a useful tool for scientific purposes.

The introduced scores for validity and reliability not only facilitate comparisons with other topic modeling methods but also serve as diagnostic tools for assessing and refining the results of the topic modeling. Furthermore, the flexibility of the Agentic RAG framework allows it to be implemented with various LLMs (e.g., LLaMA), ensuring its continued applicability as more advanced models are developed.

**Fine-Tuning LLM vs. Agentic RAG**

One may argue, that fine-tuning might be a more straight-forward solution for topic modeling with LLMs. Fine-tuning is a highly functional process of adjusting parameters of a pre-trained model for a specific task and is associated with very high computational costs for training. In a study, Mu et al. (2024b) addressed the issue of hallucinations in LLM-based topic modeling by *fine-tuning* a model using Direct Preference Optimization (DPO) technique. DPO is a streamlined version of fine-tuning without requiring extensive human annotations. This method involves pretraining the model on a subset of the dataset for a specific task before applying the fine-tuned model to the entire dataset to extract the most relevant topics. According to the authors, their approach successfully reduced the number of hallucinated topics while producing coherent and relevant topics.

We argue that Agentic RAG is a superior method for topic modeling than fine-tuning, for several reasons. First, fine-tuning an LLM requires substantial computational resources, whereas Agentic RAG operates efficiently on standard computer hardware. Second, fine-tuning is inherently a more static approach, as it depends on pretraining and may struggle with generalizing to unseen or unfamiliar data (Gao et al., 2024). Third, fine-tuning typically requires a predefined data structure for both input and output, making it less adaptable to new datasets. In contrast, RAG—particularly Agentic RAG—is more flexible, easier to evaluate, and provides greater interpretability (Gao et al., 2024). We suggest that Agentic RAG is more applicable for the task of topic modeling due to its adaptability and efficiency. Fourth, in contrast to LLM prompting or fine-tuning, Agentic RAG shows the reasoning process that the model uses to generate an output, such providing the necessary transparency for a scientific context.

**Agentic RAG for Organizational Research**

In organizational research, Agentic RAG can bridge the gap between traditional qualitative methods, such as grounded theory and content analysis, and modern LLM based approaches. Given the capability of LLMs to effectively process textual data, even in sparse contexts, they present significant potential for improving qualitative data analysis. By advancing the information retrieval process and improving contextualization, Agentic RAG expands the applicability of AI-based methods for research purposes.

Baumer et al. (2017) compared traditional grounded theory with topic modeling, concluding that each approach captures unique aspects and topics that the other might overlook. While grounded theory benefits from its expert-driven interpretive depth, topic modeling provides an objective, data-driven perspective. The integration of contextual understanding and objectivity through Agentic RAG enhances the qualitative data analysis process, delivering more accurate and contextually relevant results, that are more aligned with human understanding and offer a more comprehensive perspective on qualitative data.

**Limitations**

Despite the demonstrated advantages of Agentic RAG for topic modeling and qualitative data analysis, several limitations remain. First, our analysis utilized only a single LLM for topic generation. While the Agentic RAG framework is model-agnostic and can be implemented with any capable LLM, such as LLaMa, our evaluation was limited to GPT4-o. Given that Mu et al. (2024a) applied LLM-based topic modeling using both GPT models and LLaMa—reporting superior results with the latter—we anticipate that integrating LLaMa into Agentic RAG would yield similar or even improved outcomes. This will be addressed in future evaluations. Second, our study assessed Agentic RAG using only one dataset. While initial tests on additional datasets suggest promising results, further validation is necessary to establish the approach's generalizability. Moreover, while validity and reliability were objectively assessed through a secondary model, additional evaluations are required to further substantiate these metrics.

**Future Directions**

The integration of LLMs into research is still in its early stages, and textual data analysis provides an ideal foundation for exploring and refining their potential. Given the vast opportunities for further advancements, future research should extend this work by experimenting with different LLM agents, alternative datasets, and diverse LLM architectures. By systematically varying these key components and identifying where each approach performs best, the research process can be continuously optimized. The presented "reasoning" process by the ReAct agent is only an initial step in how LLM agents can support textual data analysis. As research on LLM agents continues to evolve rapidly, future studies should explore how these agents can further enhance the data analysis process.

Beyond text analysis, RAG also offers the potential for multimodal data analysis, integrating visual, audio, and video data. This expansion could transform data collection and interpretation across multiple research fields.

A critical aspect for future exploration is transparency. While this study has addressed the need for greater interpretability in AI-driven research, it remains essential for future work to further enhance the transparency of LLM-based methodologies, ensuring that their decision-making processes are as comprehensible as possible.

RETRIEVAL AUGMENTED GENERATION FOR TOPIC MODELING                                25**References**

Baumer, E. P. S., Mimno, D., Guha, S., Quan, E., & Gay, G. K. (2017). Comparing grounded theory and topic modeling: Extreme divergence or unlikely convergence? *Journal of the Association for Information Science and Technology*, *68*(6), 1397–1410. https://doi.org/10.1002/asi.23786

Blei, D. M., Ng, A. Y., & Jordan, M. I. (2003). Latent Dirichlet Allocation. *Journal of machine Learning research*, *3(Jan)*, 993–1022.

Chuang, J., Roberts, M. E., Stewart, B. M., Weiss, R., Tingley, D., Grimmer, J., & Heer, J. (2015). TopicCheck: Interactive Alignment for Assessing Topic Model Stability. *Proceedings of the 2015 Conference of the North American Chapter of the Association for Computational Linguistics: Human Language Technologies*, 175–184. https://doi.org/10.3115/v1/N15-1018

Duriau, V. J., Reger, R. K., & Pfarrer, M. D. (2007). A Content Analysis of the Content Analysis Literature in Organization Studies: Research Themes, Data Sources, and Methodological Refinements. *Organizational Research Methods*, *10*(1), Article 1. https://doi.org/10.1177/1094428106289252

Eisenhardt, K. M. (1989). *Building Theories from Case Study Research*.

Feuerriegel, S., Maarouf, A., Bär, D., Geissler, D., Schweisthal, J., Pröllochs, N., Robertson, C. E., Rathje, S., Hartmann, J., Mohammad, S. M., Netzer, O., Siegel, A. A., Plank, B., & Van Bavel, J. J. (2025). Using natural language processing to analyse text data in behavioural science. *Nature Reviews Psychology*, 1–16. https://doi.org/10.1038/s44159-024-00392-z

Fyffe, S., Lee, P., & Kaplan, S. (2024). "Transforming" Personality Scale Development: Illustrating the Potential of State-of-the-Art Natural Language Processing. *Organizational Research Methods*, *27*(2), 265–300. https://doi.org/10.1177/10944281231155771

**Appendix**

**Table 4**

*Glossary of specific terms*

| Terms | Meaning |
|---|---|
| Artificial Intelligence (AI) | The field of computer science that focuses on creating systems capable of performing tasks that typically require human intelligence, such as understanding natural language, recognizing patterns, and making decisions. |
| Embeddings | Numerical vector representations of words, phrases, or texts that capture their semantic meaning. Embeddings allow machines to understand and compare the meaning of words or documents based on their context and usage. |
| Large Language Model (LLM) | A type of artificial intelligence model trained on vast amounts of text data to understand and generate human language. Examples include GPT-4 and LLaMA. LLMs are capable of tasks such as text generation, summarization, and answering questions. |
| Lemmatization | A technique that groups together words that have the same inflected forms. For example, lemmatization reduces 'better' to 'good' (Feuerriegel et al., 2025). |
| LLM Agent | A specialized component within an AI system that guides the behavior of an LLM by executing specific tasks, such as retrieving relevant data, evaluating outputs, and refining prompts or queries to improve the accuracy and relevance of the model's responses. |
| LLM Fine-Tuning | The process of adapting a pre-trained LLM to a specific task or domain by training it on a smaller, specialized dataset. This customization improves the model's performance and accuracy for particular applications, such as domain-specific text generation or classification, while maintaining its general language understanding capabilities. |
| Machine Learning | A subset of artificial intelligence that involves training algorithms on data to recognize patterns and make predictions without being explicitly programmed. In text analysis, ML models are used for tasks such as classification, clustering, and topic modeling. |
| Query | A request for information or data, typically formulated as a question or a search phrase. In topic modeling and retrieval systems, a query is used to retrieve relevant documents or information from a knowledge base. |
| Stemming | A technique similar to lemmatization used to reduce words to their base or root form. For instance, stemming reduces the words 'fishing', 'fished' and 'fisher' to 'fish'. Stemming is a common feature reduction technique (Feuerriegel et al., 2025). |
| Stop Word removal | The process of removing common words (e.g., 'and', 'the' and 'is') that appear frequently in texts but offer little value in understanding the meaning. Stop word removal reduces noise (Feuerriegel et al., 2025). |

Note. Terms appearing in italics throughout the text are defined in this glossary.

**Table 5**

*Topics generated with LLM prompting and LDA*

| LLM prompting (Mu et al. 2024a) | LDA (Mu et al. 2024a) |
| --- | --- |
| Safety & Side Effects | Vaccine Safety |
| Medical & Health Concerns | COVID-19 Vaccination |
| Trust & Misinformation | Vaccine Safety |
| Political & Social Influences | COVID vaccine |
| Economic & Corporate Concerns | Vaccine effectiveness |
| Efficacy Doubts | COVID-19 Vaccine |
| Autonomy & Personal Beliefs | Vaccine Allergies |
| Mandatory Vaccination Concerns | COVID-19 Vaccine |
| Access & Availability | COVID-19 Vaccination |
| Others | COVID-19 Vaccination |